\documentclass[conference]{IEEEtran}
\usepackage[left=1.57cm,right=1.57cm,top=0.95cm,bottom=2.54cm]{geometry}

\IEEEoverridecommandlockouts

\usepackage{booktabs}
\usepackage{multirow}
\usepackage{algorithm}
\usepackage{algpseudocode}
\usepackage{enumitem}

\usepackage{url}
\usepackage{cite}
\usepackage{amsmath,amssymb,amsfonts}
\usepackage{graphicx}
\usepackage{textcomp}
\usepackage{xcolor}
\def\BibTeX{{\rm B\kern-.05em{\sc i\kern-.025em b}\kern-.08em
    T\kern-.1667em\lower.7ex\hbox{E}\kern-.125emX}}

\usepackage{float}
    
\begin{document}
%
\title {Driver Drowsiness Detection Model Using Convolutional Neural Networks Techniques for Android Application}

%
%
%


\author{
	\IEEEauthorblockN{
	Rateb Jabbar\IEEEauthorrefmark{1}\IEEEauthorrefmark{2},
	Mohammed Shinoy\IEEEauthorrefmark{1}, 
	Mohamed Kharbeche\IEEEauthorrefmark{1}, 
	\\
	Khalifa Al-Khalifa\IEEEauthorrefmark{4},
	Moez Krichen\IEEEauthorrefmark{3},
	Kamel Barkaoui\IEEEauthorrefmark{2}
	}

	\\
	\IEEEauthorblockA{
	\textit{\IEEEauthorrefmark{1}Qatar Transportation and Traffic Safety Center, Qatar University, Qatar} \\
	{\{rateb.jabbar,m.shinoy,mkharbec\}}@qu.edu.qa \\
	}
	\\
	\IEEEauthorblockA{
	\textit{\IEEEauthorrefmark{4}College of the North Atlantic-Qatar, Doha, Qatar} \\
	alkhalifa@cna-qatar.edu.qa \\
	}
	\\
	\IEEEauthorblockA{
	    \textit{\IEEEauthorrefmark{2}Cedric Lab, Computer Science Department,Conservatoire National des Arts et Meteirs, France } \\
	    {\{jabbar.rateb.auditeur,kamel.barkaoui\}}@cnam.fr
	}
	\\
	\IEEEauthorblockA{
	    \textit{\IEEEauthorrefmark{3}ReDCAD Laboratory, National School of Engineers of Sfax, University of Sfax, Tunisia }\\
		{moez.krichen}@redcad.org \\
    } 
}

\maketitle

\begin{abstract}


A sleepy driver is arguably much more dangerous on the road than the one who is speeding as he is a victim of microsleeps. Automotive researchers and manufacturers are trying to curb this problem with several technological solutions that will avert such a crisis. This article focuses on the detection of such micro sleep and drowsiness using neural network-based methodologies. Our previous work in this field involved using machine learning with multi-layer perceptron to detect the same. In this paper, accuracy was increased by utilizing facial landmarks which are detected by the camera and that is passed to a Convolutional Neural Network (CNN) to classify drowsiness. The achievement with this work is the capability to provide a lightweight alternative to heavier classification models with more than 88\% for the category without glasses, more than 85\% for the category night without glasses. On average, more than 83\% of accuracy was achieved in all categories. Moreover, as for model size, complexity and storage, there is a marked reduction in the new proposed model in comparison to the benchmark model where the maximum size is 75 KB. The proposed CNN based model can be used to build a real-time driver drowsiness detection system for embedded systems and Android devices with high accuracy and ease of use.
\end{abstract}


%
\IEEEpeerreviewmaketitle

\section{Introduction}\label{sec:01}

The National Highway Traffic Safety Administration (NHTSA) statistics show that about 2.5\% of all fatalities during a crash is attributed to drowsy driving \cite{HighwayTrafficSafetyAdministration2017}. In 2015 alone, the total number of crashes related to drowsy driving was over 72,000. Remarkably, vehicle crashes caused by impaired driving are much more frequent than those caused by intoxicated drivers \cite{walker2017we}. People who drive under the influence are just slower in their reaction time when compared to normal active drivers. Drowsy driving, on the other hand, makes the drivers a victim of microsleeps.

As nearly 1.4 million people die on the road every year thereby making it the 7th global cause of death in 2016 \cite{who}, it is no wonder that the automotive industries, research institutions, and other governmental agencies are creating technologies that prevent such scenarios. Among the many crash aversion techniques and other safety features that are developed in newer cars, there is a lot of interest on detecting fatigued and drowsy drivers i.e by using cameras, sensors and other instruments to alert and thereby avert fatal crashes.

Automotive companies such as Mercedes-Benz \cite{mercedes}, Tesla \cite{mercedes}, and others have their forms of driver assistance systems. Such as variable cruise control, automatic braking systems, lane departure warnings and assisted steering. These technologies have helped drivers to stop crashes from happening. Recently Samsung has partnered with Eyesight \cite{eyesight} to track a drivers attention by reading facial features and patterns. Such a tracking system can warn drivers how attentive they are while driving. Although these advancements are there most of these systems are proprietary and limited to high end vehicles.

There has been a significant rise in the number of vehicles that are equipped with Android Auto or Apple Car nowadays. Most of the cars that are being introduced now have these components built-in. Such features are now readily available in lower-end cars too. Due to this, drowsiness detection systems can be easily developed around such built-in Android and iOS platforms. Embedded devices or mobile phones that can easily pair with these car dashboards can be used to enhance driver behavior detection using simple camera setup and state of the art computer vision systems powered by Deep Learning.

Microsleeps are of short duration where the driver has his eyes closed and is not perceiving any visual information and thereby cannot react if the car departs from the lane or if the car comes to halt due to harsh braking. With the advent of new sensors and radars in high-end cars, cars can notify or even take action to avoid such crashes. Although this is a great advancement, it would also be beneficial if the car knows that the driver is sleepy and requests him to take rest. Also, another common issue with the current machine vision models is the fact that most of these algorithms are bulky ( large in size) and require dedicated hardware to run the models that have been developed. They do not run efficiently on devices with low computational power. This paper details such a Deep Learning-based drowsy detection algorithm that can potentially enable such an intervention, Moreover, it is simple and easy to configure on any mobile or embedded device. 

The rest of the paper is divided as follows. In the next section \ref{sec:02}, the literature review, a few of the state-of-the-art systems for driver drowsiness detection systems are presented. In section Solution and Methodology \ref{sec:03}, the proposed algorithm and the methodology based on CNN and Facial Landmark Detection (D2CNN-FLD) are described. A discussion regarding the implementation of an Android system with this algorithm is also explained. Furthermore, the accuracy, efficiency, and comparisons of the given model are detailed in Experimental Results in \ref{sec:04}. Finally, conclusions \ref{sec:05} and possible avenues to be investigated in the future are presented.

\section{Literature review}\label{sec:02}

As sleep-induced crashes represent a significant portion of vehicle crashes in the world, researchers and automotive companies have created different solutions ranging from finding patterns in driving habits to analyzing brain waves and vitals of the driver while driving. Most of these solutions are backed by some predictive algorithms powered by statistics and machine learning. The most common ones can be broadly divided into three categories as described in the following paragraphs. 

One approach is to find changes in vehicle behavior, such as the one proposed by McDonald et al. ~\cite{McDonald2018}. He created a contextual and temporal algorithm that utilizes the steering angle, vehicle speeds, and accelerator pedal positions. These values are passed into a Bayesian Network which determines if a driver shows characteristics of drowsy behavior. The algorithm was found to have lower false-positive rates than PERCLOS \cite{perclos} methods, which predicts drowsiness based on eyelid movements and patterns. The takeaway from this study was that to predict correctly, the context of the situation is crucial. The data that it captures over a previous 10-second period is vital in understanding whether the person is at risk of drowsiness related lane departures.

A second approach is based on studies focused on using the drivers' vitals, brain waves, and readings from Electro-Encephalo Grams (EEG's) to make predictions. Wei et al. \cite{Wei2018} made comparisons between non-hair bearing EEG Brain-Computer Interfaces that are easy to wear and less intrusive than the lab-based whole scalp EEG's which are less comfortable. The study showed that non-hair bearing devices had no significant reduction in performance when compared to whole-scalp EEG. Thereby with this finding one may develop less intrusive and comfortable headbands. EEG alone is unable to detect all stages of drowsiness, so Kartsch et al. \cite{Kartsch2018} used EEG with Inertial Measurements Units (IMU) sensors to detect 5 levels of drowsiness with about 95\% accuracy. The team fused behavioral information from IMU and EEG information to detect drowsiness. Another drawback of the EEG system was the power requirements of these devices. Their technology has also facilitated the implementation of a parallel ultra-low power (PULP) platform on a microcontroller which extended the battery life to almost 46 hours, thereby creating devices that are always wearable and require low maintenance. Tateno et al. \cite{Tateno2018} developed a system that just uses heart rate monitoring to detect the respiration of a person and thereby detect drowsiness. The methodology was found to be an effective predictor for respiration and thereby drowsiness.

Yet another technique is to utilize the power of computer vision. The recent breakthroughs in Deep Learning have provided new tools to computer vision for detection and classification. Computer vision-related applications utilize these methods in object detection, health and wellness, and even agricultural applications too \cite{kamilaris2018}. A big impact in this space has been for imaging data. Since drivers’ facial features change significantly once he gets tired, computer vision scientists have attempted to capitalize on this and use it to provide solutions for drowsiness detection. Tayab Khan et al. \cite{TayabKhan2019} proposed a solution to measure the angle of eyelid curvature and thereby identify if the eyes are closed or not. They achieved 95\% accuracy with this method, but the limitation is that there needs to be enough light for the method to work as it functions poorly at night. Shakeel et al. \cite{Shakeel2019} used MobileNet-SSD architecture to train a custom dataset of 350 images. The model was capable of achieving a Mean Average Precision of 0.84. The system was cost-effective and efficient as the algorithm could be deployed in an Android device and the camera stream could be classified in real-time. Celona et al. \cite{Celona2018} proposed a vision-based Multi-Task Driver Monitoring Framework that analyzes eyes, mouth, and the pose of the head simultaneously to predict the level of the drowsiness. This study was conducted on the NTHU \cite{weng2016driver} dataset. Another study conducted by Xie et al. \cite{Xie2019} used transfer learning and sequential learning from yawning video clips to detect yawning on the YawDD and NTHU-DDD database. This system was able to have higher precision and was robust to changes in the position and angle of the face to the camera. Mehta et al. \cite{Mehta2019} developed an Android application that is capable of detecting facial landmarks and then computing the Eye Aspect Ratio (EAR) and Eye Closure Ratio (ECR) to predict driver’s drowsiness based on machine learning models with an accuracy of 84\%.

A method that many companies are trying to follow is to combine the three different techniques explained above and utilize multiple inputs to come to a decision. A start-up Ellcie-Healthy \cite{ellcie} has developed a smart glass that embeds a drowsiness detection application by including blink detection, eye tracking, and vitals monitoring. The smart glass monitors these inputs and provides drowsiness intervention by beeping and thereby asking the driver to take a rest. The combination techniques require multiple sensors such as infrared, cameras and heart rate monitors on a single platform to provide great results. However, these instruments are quite expensive and require proprietary solutions to be set up.

All these innovations in deep learning space come at a cost, a bigger model with higher computational requirements. This is the bane of deep learning. In our previous work \cite{jabbar2018} \cite{jabbar2018qfarc} a machine learning model that classified drowsy images using a multi-layer perceptron-based model [D2MLP-FLD] was proposed. The result was a respectable accuracy of 81\% while creating a model that was 100KB in storage size. The idea that is proposed in this paper is to improve upon the algorithm and create a lightweight application in terms of performance as well as storage space.

\section{Solution and Methodology}\label{sec:03}
\subsection{Dataset and Preprocessing}

The Deep Learning model developed here is trained on videos obtained from the National Tsing Hua University (NTHU) Driver Drowsiness Detection Dataset \cite{weng2016driver}. These videos are preprocessed to create images for this study.

\subsubsection{NTHU Dataset}

\begin{figure}[!t]
	\centering
	\includegraphics[width=.5\textwidth]{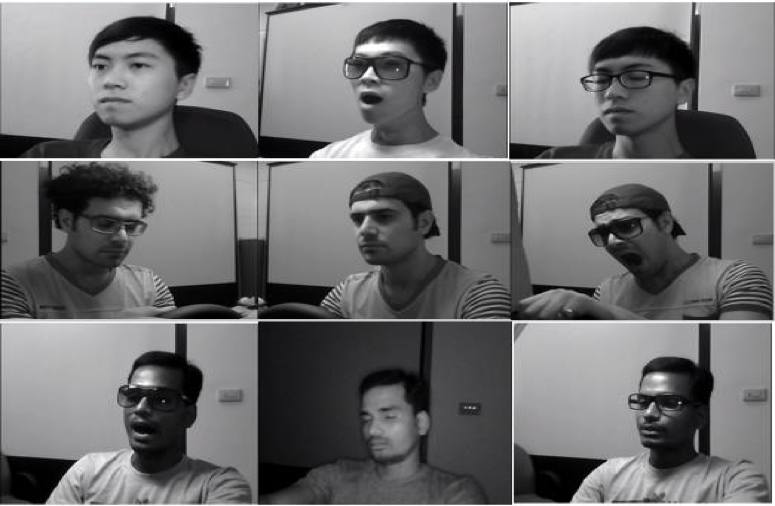}
	\caption{NTHU Dataset Sample Images}
	\label{fig:sample_dataset}
\end{figure}

 This dataset contains 22 subjects from different ethnicities who have been recorded under day and night conditions. The driving situations and behavior were simulated, which included conversations, yawning, slow rate blinking, sleepy head movements, and drowsy eyes as shown in \ref{fig:sample_dataset}. The recording was done using infrared cameras to obtain night-time videos too. The resulting content is 9.5 hours of videos that have a resolution of 640*480 at 30 frames per second containing different subjects enacting regular and drowsy driving behaviors.
 
 \subsubsection{Preprocessing the dataset}

Video prepossessing happens in 4 stages as shown in the Figure.~\ref{fig:video_processing}:

\begin{figure}[!t]
	\centering
	\includegraphics[width=.5\textwidth]{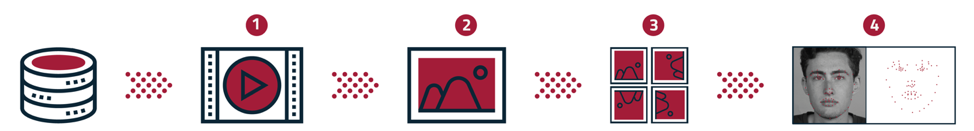}
	\caption{Video preprocessing outline}
	\label{fig:video_processing}
\end{figure}

\begin{itemize}
	\item Step 1 – Extracting Images from Video Frames:
	
	As we cannot train the neural network on video data, the images need to be extracted and then fed into the model. The video specifications are at 30 frames per second. These frames are converted to images and thereby correspond to a total of around 600,000 images.
	
	\item Step 2 – Data Augmentation:
	
	In the Deep Learning training phase, it is important to have more data so that the model can learn all the nuances and variations in the images. A common method to increase the number of training points is to use data augmentation. Codebox \cite{codebox} was used to generate new images by performing a set of augmentation operations on the images extracted from video frames, as illustrated in Figure \ref{fig:data_augmentation}.

\begin{figure}[!t]
	\centering
	\includegraphics[width=.5\textwidth]{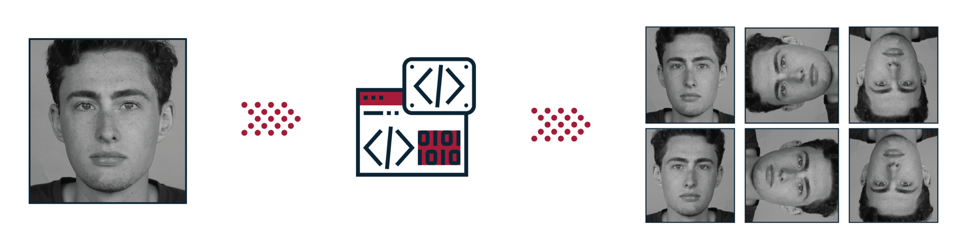}
	\caption{Data Augmentation}
	\label{fig:data_augmentation}
\end{figure}

\item Step 3 – Extracting landmark coordinates from images: 

In order to locate and represent salient parts of the face such as eyes, mouth, nose and the jawline, facial landmarks are used. These landmarks are essential for head pose estimation, blink detection, yawning detection, etc. An open-source C++ library, Dlib \cite{dlib} has pre-written functions that are used to obtain facial landmarks. This library is programmed to find the x, y coordinates of 68 facial landmarks to map the facial structure as shown in (Figure \ref{fig:facial_feature_detection}).

Dlib Library utilizes the power of OpenCV’s Haar Cascades to detect facial landmarks. Paul Viola and Michael Jones ~\cite{viola2001} introduced this technique which was optimized over the years as an Open Source initiative by OpenCV library contributors based on machine learning. This technique uses many images that are either positively or negatively marked for the presence of an object that is to be detected. In our case, this object would be a face. From this, the algorithm will be able to detect new facial images that are fed into the algorithm.

\begin{figure}[!ht]
	\centering
	\includegraphics[width=.5\textwidth]{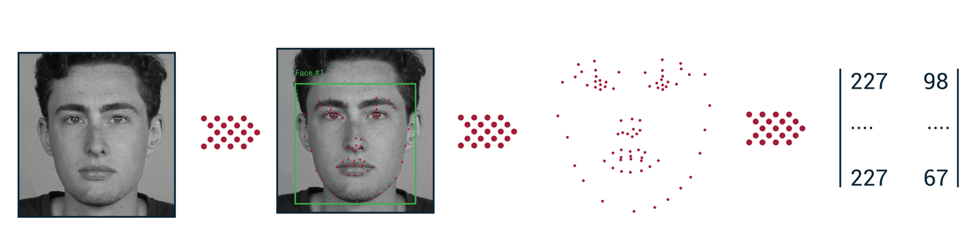}
	\caption{Facial feature detection}
	\label{fig:facial_feature_detection}
\end{figure}

\end{itemize}

\subsection{Drowsiness Detection based on CNN and Facial Landmark Detection (D2CNN-FLD)}

The input to the algorithm is the facial landmark coordinates which are extracted from the images using the DLib Library as represented in the algorithm \ref{algo:02}. The presented model is based on the CNN approach with five layers. The first four convolution layers include a Linear \cite{mlp} activation function with Leaky ReLU \cite{maas2013} function to enable a small gradient when the unit is not active.
\begin{equation}
f(x)=\left\{
\begin{array}{ll}
x, & \hbox{If $x>0$;} \\
0.1x , & \hbox{otherwise.}
\end{array}
\right.
\end{equation}
Furthermore, MaxPooling \cite{Goodfellow-et-al-2016} function is used to compute the maximum value from each cluster of neurons at the prior layer. In order to prevent overfitting, a Dropout function is used \cite{mlp} and the dropout rate is set at 25.4\% after testing out a range between 20 - 30. The output layer is the output of a Softmax \cite{Goodfellow-et-al-2016} function to get output class label probabilities. The detailed algorithm of the D2CNN-FLD is described below.

\begin{algorithm}[!t]
	\caption{Real-Time Driver Drowsiness Detection Algorithm with CNN}
	\textbf{Input:} DLib Facial landmark positions and labels\\
	\label{algo:02}
	\textbf{Output:} 2 - Class probabilities of drowsy driving
	\begin{algorithmic}[1]
		\State Loading the Data 
		\State Min-Max Scaling to normalize the dataset between the range 0 and 1.
		\State Adding hidden layers and dropout:
		\Statex a.~The first convolutional layer with linear activation. The input layer consists of 67*2=134 nodes with a total of 100 neurons.
		\Statex b.~A LeakyReLU function allows a small gradient when the unit is not active. The alpha rate is set to 10\%.
		\Statex c.~A MaxPooling function to reduce the number of parameters within the model
		\Statex d.~ Dropout is set to 25.4\% to prevent over-fitting.
		\Statex e.~For j=1 to 3 do
		\begin{itemize}[leftmargin=.35in]
			\item[(i).] Adding convolution hidden layer with a linear function. The number of neurons used is 1024.
			\item[(ii).] A LeakyReLU function allows a small gradient when the unit is not active. The alpha rate is set to 10\%.
			\item[(iii).] A MaxPooling function to reduce the number of parameters within the model0
			\item[(iv).] Dropout is set to 20\% to prevent over-fitting.
		\end{itemize}
		\Statex f. Final fully connected layer with a softmax function to get a 2 class output probabilities.
		\State The model is trained until acceptable accuracy.
	\end{algorithmic}
\end{algorithm}
\subsection{Android Implementation and Architecture}

An integral part of the whole process is the application that is used to capture the image data and send it for processing. The Android mobile application is capable of taking pictures of the driver. This image is analyzed by the Dlib library. Then, the Java Native Interface (JNI) \cite{java} transmits the image data from the native Android application which is programmed in Java and the Dlib library which is developed in C++.

\begin{figure}[!t]
	\centering
	\includegraphics[width=.5\textwidth]{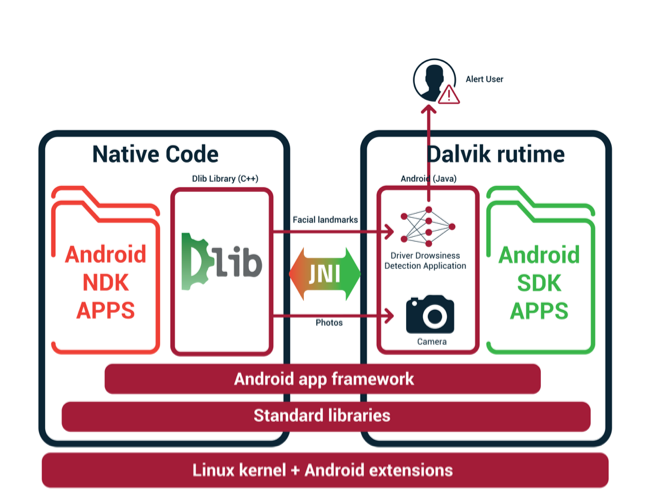}
	\caption{Proposed architecture of solution in Android.}
	\label{fig:android_app_architecture}
\end{figure}

As illustrated in figure \ref{fig:android_app_architecture}, the Dlib library on receiving the image data, will do the pre-processing and extract the facial landmarks and send the data to the above developed CNN model. This data runs through the neural network and the algorithm evaluates if the driver is drowsy or not. The results of the images are indicated on the application in a real-time fashion; if the driver is found to be drowsy, the application will send notification via visual and audio messages.
\section{Experimental results}\label{sec:04}
As detailed in Table \ref{tab:01}, a total of 18 subjects having sleepy and non-sleepy status for each subject were selected for each scenario from the dataset to form the training dataset. In addition, 4  subjects were included for the evaluation. The dataset consists of 5 different simulated driving scenarios namely with glasses, with sunglasses, without glasses, night with glasses, and night without glasses. In total, 200 videos (30fps) were used and converted to images. Codebox Algorithm is used to generate new images based on the NTHU images. furthermore, the coordinates of the facial landmark were extracted from the generated images. The number of images and videos used for each state is shown in Table \ref{tab:01}.

\begin{table*}[ht]
	\centering
	\caption{Image Categories}
	\label{tab:01}
	\begin{tabular}{@{}llcccc@{}}
		\toprule
		Dataset & Category & Nbr. Videos & \begin{tabular}[c]{@{}c@{}}Nbr. Images\\ Extracted\end{tabular} & \begin{tabular}[c]{@{}c@{}}Nbr. Images detected\\ by Dlib\end{tabular} & \begin{tabular}[c]{@{}c@{}}Nbr. Images used\\ After Data Augmentation\end{tabular} \\ \midrule
		Training & With glasses & 36 & 106,882 & 93,301 & 559,806 \\
		& Night Without glasses & 36 & 52,372 & 43,659 & 261,954 \\
		& Night With glasses & 36 & 50,991 & 39,959 & 239,754 \\
		& Without glasses & 36 & 108,380 & 96,037 & 576,222 \\
		& With sunglasses & 36 & 107,990 & 83,716 & 502,296 \\ \midrule
		Evaluation & With glasses & 4 & 37,357 & 34,879 & 209,274 \\
		& Night Without glasses & 4 & 29,781 & 27,322 & 163,932 \\
		& Night With glasses & 4 & 32,922 & 31,533 & 189,198 \\
		& Without glasses & 4 & 45,005 & 40,714 & 244,284 \\
		& With sunglasses & 4 & 28,214 & 25,479 & 152,874 \\ \midrule
		Total &  & 200 & 599,894 & 512,308 & 3,073,848 \\ \bottomrule
	\end{tabular}
\end{table*}

\begin{table*}[ht]
	\centering
	\caption{Summary of experimental results. (‘-’ means ‘not applicable’)}
	\label{tab:04}
	\begin{tabular}{llcccc}
		\hline
		Dataset & Category & \begin{tabular}[c]{@{}c@{}}D2MLP-FLD\\ MLP-Model\end{tabular} & \begin{tabular}[c]{@{}c@{}}D2CNN-FLD\\ CNN Model\end{tabular} & \begin{tabular}[c]{@{}c@{}}Faster RCNN\\ (VGG-16)\end{tabular} & \begin{tabular}[c]{@{}c@{}}Faster RCNN\\ (AlexNet)\end{tabular} \\ \hline
		Accuracy (\%) & Validation & 80.9 & 83.3 & 90.5 & 82.8 \\ \cline{2-6} 
		\hline
		\multirow{2}{*}{Compression (MB)} & Model size & 0.1 & 0.075 & 547 & 236 \\ \cline{2-6} 
		& GPU Memory & 44.1 & 38.9 & 3183 & 845 \\ \hline
		\multirow{2}{*}{\begin{tabular}[c]{@{}l@{}}Drowsiness\\ Detection time (ms)\end{tabular}} & \begin{tabular}[c]{@{}l@{}}Dell workstation\\ NVIDIA Quadro P4000\end{tabular} & 41.6 & 38.45 & - & - \\ \cline{2-6} 
		& Samsung Galaxy S8 plus & 154 & 142 & - & - \\ \hline
		\multirow{2}{*}{\begin{tabular}[c]{@{}l@{}}Overall speed\\ (fps)\end{tabular}} & \begin{tabular}[c]{@{}l@{}}Dell workstation\\ NVIDIA Quadro P4000\end{tabular} & 7.4 & 6.87 & 9.1 (GTX-1080) & 22 (GTX-1080) \\ \cline{2-6} 
		& Samsung Galaxy S8 plus & 267.56 & 234.25 & - & - \\ \hline
	\end{tabular}
\end{table*}
The experiments were conducted on a Dell workstation with Intel Xeon Gold 6128, 3.4 GHz, 64 GB RAM, and NVIDIA QUADRO P4000. The accuracy results per driving scenarios using the new technique D2CNN-FLD method are shown in Table \ref{tab:03}; its overall accuracy was 83.3\%. This is a marked 2.5\% improvement in the results when compared to the MLP method that was utilized in the prior study.


\begin{table}[ht]
	\centering
	\caption{Accuracy per driving using CNN method}
	\label{tab:03}
	\begin{tabular}{@{}lc@{}}
		\toprule
		Category & Accuracy(\%) \\ \midrule
		With glasses & 83.76 \\
		Night Without glasses & 85.82 \\
		Night With glasses & 79.45 \\
		Without glasses & 88.89 \\
		With sunglasses & 78.72 \\
		All & 83.33 \\ \bottomrule
	\end{tabular}
\end{table}
Evaluation and experimentation convey clearly that the eyes are a significant factor for drowsiness classification in any circumstance. In images where the eyes are obstructed by wearing sunglasses, the classification accuracy drops a few percentages. Another factor that was found to affect the classification accuracy is the illuminance on the subject's face which when increased tend to display the features on the face much more clearly. Images that lack clarity due to uneven lighting or darkness is an important performance factor since the error rate increases by 3\% using D2CNN-FLD. 

The main concept and contribution behind this work is to create models that are low in calculations capacity requirement so that it can be utilized in embedded systems. The experimental results show that the accuracy averages at 83.33\% using the D2CNN-FLD method while maintaining a very small model size.

Experiments were performed using the model D2CNN-FLD. D2MLP-FLD \cite{jabbar2018}, Faster RCNN \cite{Ren2017} with VGG-16\cite{Simonyan2015} and AlexNet \cite{alex2012} models are used as benchmarks for comparative purposes. A summary of the overall experimental results is shown in Table \ref{tab:04}. The models were compared in terms of accuracy, compression, drowsiness detection time, and the overall speed. The D2CNN-FLD outperformed D2MLP-FLD in both validation and test categories of accuracy. Regarding the size of the model, D2CNN-FLD is 0.075 MB that is 25\% improvement over the older D2MLP-FLD which is about 0.1 MB.

Similarly, D2MLP-FLD required greater GPU computational memory (44.1MB) more than D2MLP-CNN. The two models also differ in drowsiness detection time (ms). When NVIDIA Quadro P4000 is used, in terms of time, D2MLP-FLD achieved 41.6 ms while D2CNN-FLD achieved 38.45 ms. The results are different for Samsung Galaxy S8 plus, D2MLP-FLD and D2CNN-FLD achieved, respectively, 154ms and 142ms. Moreover, the MLP achieved an overall speed of 7.4fps with NVIDIA Quadro P4000 and 267.56 fps for Samsung Galaxy S8 plus, whereas D2CNN-FLD achieved 6.87 fps and 234.25fps, respectively.

As for model size, complexity and storage, there is a marked reduction in the new proposed model D2CNN-FLD in comparison to the older D2MLP-FLD model. As pointed out, the maximum size of the developed D2CNN-FLD model is 75 KB. On the contrary, the state-of-the-art VGG16 Based RCNN13 model comes up to 547 MB of disk space, these models require much higher RAM and computational resources. AlexNet based faster-RCNN models are equal to 236 MB, which is significantly much larger in terms of storage capacity. Although there have been improvements made in compressing these models, they still take up approximately 12 MB \cite{han2015deep}.

Furthermore, Both D2CNN-FLD and D2MLP-FLD require only 44.1 MB and 38.9 MB of RAM requirement during computation while faster-RCNN need more than 3180 MB of RAM. Hence, owing to its small model size and computing memory needs, it is possible to integrate the developed models in smaller embedded systems.

\section{Conclusion}\label{sec:05}

The paper described an improved drowsiness detection system based on CNN-based Machine Learning. The main objective is to render a system that is lightweight to be implemented in embedded systems while maintaining and achieving high performance. The system was able to detect facial landmarks from images captured on a mobile device and pass it to a CNN-based trained Deep Learning model to detect drowsy driving behavior. The achievement here was the production of a deep learning model that is small in size but relatively high in accuracy. The model that is presented here has achieved an average of 83.33\% of accuracy for all categories where the maximum size of the model did not exceed 75KB. This system can be integrated easily into dashboards in the next generation of cars to support advanced driver-assistance programs or even a mobile device to provide intervention when drivers are sleepy. There are limitations to this technology, such as obstructing the view of facial features by wearing sunglasses and bad lighting conditions. However, given the current state, there is still room for performance improvement and better facial feature detection even in bad lighting conditions.



%

%

\section*{Acknowledgment}

This publication was made possible by an NPRP award [NPRP8-910-2-387] from the Qatar National Research Fund (a member of Qatar Foundation). The statements made herein are solely the responsibility of the authors.

\ifCLASSOPTIONcaptionsoff
  \newpage
\fi



%

\bibliographystyle{IEEEtran}
\bibliography{paper}

\end{document}